\documentclass{article}

\usepackage[final,dblblindworkshop]{neurips_2026}

\usepackage[utf8]{inputenc}
\usepackage[T1]{fontenc}
\usepackage{url}
\usepackage{booktabs}
\usepackage{amsfonts}
\usepackage{amsmath}
\usepackage{nicefrac}
\usepackage{microtype}
\usepackage{graphicx}
\usepackage{multirow}
\usepackage{xcolor}
\usepackage{colortbl}
\usepackage{wrapfig}
\usepackage{hyperref}

\usepackage{algorithm}
\usepackage{algpseudocode}

\definecolor{goldrow}{HTML}{F6E7A8}
\definecolor{silverrow}{HTML}{E3E3E6}
\definecolor{bronzerow}{HTML}{EBD3B6}

\newcommand{\finding}[1]{\noindent\textbf{$\triangleright$ #1}\;}

\title{CIPHER: A decoupled exploration-selection framework for test-time scaling of data science agents}

\author{Maxime Heuillet \quad Sharadind Peddiraju \\ Amazon Web Services \\ mheuille@amazon.com}

\begin{document}

\maketitle

\begin{abstract}
Enterprise data science and analytics spans closed-ended information extraction with verifiable answers and open-ended insight discovery. Agents are evaluated on one or the other, so there is no evidence on whether one system can serve both. We present \texttt{CIPHER}, a data science agent built on a Decoupled Exploration--Selection (DES) framework that applies test-time scaling at the plan level: generating $N$ candidate analysis plans, selecting $M{<}N$ for parallel execution, and aggregating the results. We evaluate $48$ distinct design choices for the DES framework (spanning generation, selection, budget and aggregator capacity) jointly on $357$ tasks from Infi-DA-Bench (closed-ended) and InsightBench (open-ended), on a proprietary and an open-weights base model. A single zero-tuning configuration is competitive against purpose-built specialist baselines under matched models, within $1.2$pp of \texttt{DataWise} and $1.3$pp of \texttt{Agent-Poirot}. The joint sweep exposes that generation and aggregation are the most influential performance levers, while random selection --- the only strategy with zero overhead --- matches its tuned alternatives. These results offer new insights on the design of test-time scaling strategies, and on how a fixed inference budget is best allocated across them, for agents deployed on enterprise analytics workloads.
\end{abstract}

\section{Introduction}

We focus on enterprise data analytics, which increasingly relies on AI agents that operate over business data with minimal oversight --- and that must do so at an inference cost that deployment can sustain. Data science tasks are heterogeneous: the same agent must resolve closed-ended tasks with verifiable answers \citep{hu2024infiagent}, and open-ended tasks whose goal is to surface actionable insights \citep{sahu2025insightbenchevaluatingbusinessanalytics}. 

Existing data analytics agents such as \texttt{DataWise} \citep{you2025datawiseagentnotebookcentricllmagent} and \texttt{Agent-Poirot} \citep{sahu2025insightbenchevaluatingbusinessanalytics} instantiate a single execution route conditioned on one initial plan, making them vulnerable to cascading errors from a suboptimal start. 
In contrast with these prior works, we present \texttt{CIPHER}, an agent that applies test-time scaling \citep{brown2024largelanguagemonkeysscaling, zhang2025optimizingsequentialmultisteptasks} at the plan level: it generates $N$ candidate plans, selects $M{<}N$ for parallel execution, and aggregates the outcomes. Appendix~\ref{sec:related_works} positions this design in the broader test-time scaling literature. 

Our contributions are: 
(1) we present CIPHER, an agent able to resolve closed-ended and open-ended enterprise analytics tasks;
(2) through \texttt{CIPHER}, we introduce a Decoupled Exploration-Selection (DES) framework which is a novel approach for test-time scaling based on the generation followed by the selection of initial states, opening the way for future research on this problem.
(3) We propose an extensive empirical study on the optimal design of the DES framework, identifying which design axes convert additional inference compute into accuracy and which consume budget without measurable return. The evaluation focuses on enterprise data analytics, and to our knowledge it is the first evaluation on both closed- and open-ended tasks under matched models, on a proprietary and an open-weights base model.

\begin{wrapfigure}[23]{r}{0.5\textwidth}
    \vspace{-15pt}
    \centering
    \includegraphics[width=\linewidth]{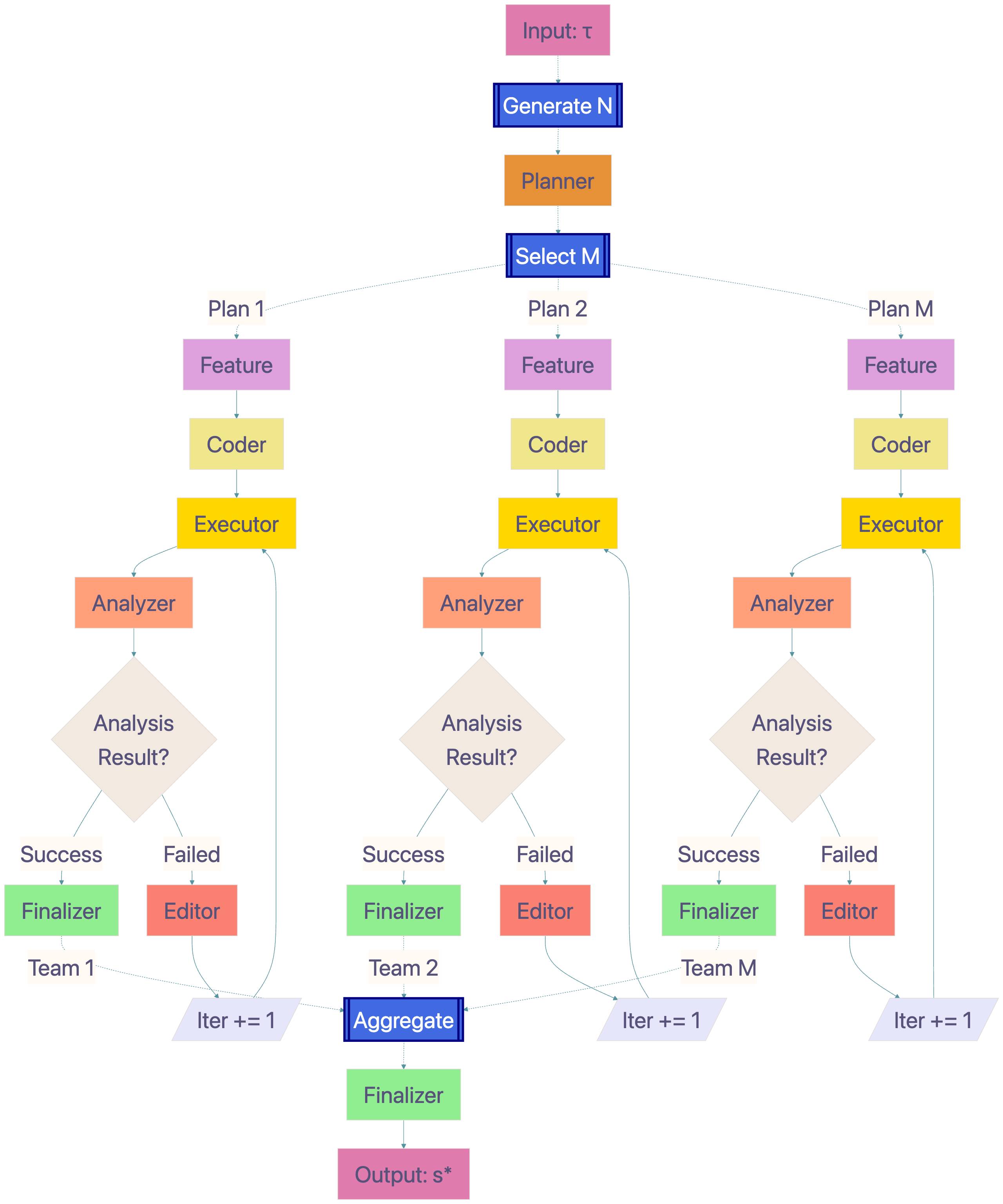}
    \caption{\texttt{CIPHER} execution route. }
    \label{fig:arch}
\end{wrapfigure}

\section{CIPHER: a data science agent}

\texttt{CIPHER}'s execution route is illustrated in Figure \ref{fig:arch}.
The agent receives as input a task noted $\mathcal{T}$ composed of a dataset and a textual user goal (open-ended or closed ended data science question). 
Examples of tasks and executions are provided in Figures 4 and 5 of the Appendix.
The execution route consists of a deterministic directed graph implemented in LangGraph \citep{langgraph2024}. The output is a textual response to the user goal noted $s^*$. 

\textbf{A novel approach for initial state generation and selection} 
\texttt{CIPHER} first generates a broad set of $N$ candidate states, where $N$ is a hyperparameter to set by the user. The candidates are generated by a \texttt{Planner} agent instructed to create detailed and actionable plans aiming to solve the task. Subsequently, a subset of $M<N$ states is selected for execution ($M$ is a hyperparameter set by the user as well). Together, parameters $M$ and $N$ control the novel Decoupled Exploration-Selection (DES) framework further detailed in Section \ref{sec:framework}.
This framework conceptually differs from existing works on test-time scaling, which typically conflate the two steps as a finite set of $M$ states is typically generated and is then fully executed \citep{zhang2025optimizingsequentialmultisteptasks,zhu2025scalingtesttimecomputellm}. 

\textbf{Parallel routes execution}
\texttt{CIPHER} executes the $M$ selected states in parallel to decrease latency. 
Each selected state follows the same execution route detailed in Figure \ref{fig:arch}. The route is composed of a \texttt{Feature} agent (suggests the creation of relevant features), a coding loop (translates the plan into executable code), and a \texttt{Finalizer} agent (analyze code executions and provide a summary). Finally, an \texttt{Aggregator} agent synthesizes the information from the concurrent routes, enabling the final output.

\textbf{Code iterative refinement Loop}
\texttt{CIPHER} instantiates an iterative refinement loop \citep{chen2023teaching} to improve the quality of the generated code. A \texttt{Coder} agent generates a first code draft, which is then executed. Then, an \texttt{Analyzer} agent evaluates jointly the logs and the code to identify errors or suboptimal patterns. If issues are detected, an \texttt{Editor} agent modifies locally the code draft without re-generating it all. The loop repeats until the iteration budget $b$ is reached ($b$ is a hyper-parameter set by the user). We provide the  descriptive statistics on the effectiveness of the coding loop in Figure \ref{fig:coding_loop} in the Appendix.

\section{The Decoupled Exploration-Selection (DES) Framework}
\label{sec:framework}

The Decoupled Exploration-Selection (DES) framework draws structural inspiration from the classic exploration-exploitation trade-off in decision theory and reinforcement learning \citep{lattimore2020bandit}. By separating candidate generation ($N$) from state execution ($M$), it maps wide-breadth search space exploration, to a distinct phase of strategic, value-maximizing exploitation. 

To our knowledge, explicitly decoupling candidate generation from strategic selection is a novel design point in the test-time scaling design space, distinct from prior work that couples the two steps \citep{inoue2025widerdeeperscalingllm,wang2025thinkdeepthinkfast}. 
The integration of the framework with \texttt{CIPHER} is illustrated in Figure \ref{fig:arch}, and is further detailed in Algorithm \ref{alg:framework}.

\subsection{Generating candidate states}

The method \textsc{Generate-Candidates} in Algorithm \ref{alg:framework} outputs a set $\mathcal C$ (with $|\mathcal C|=N$) of candidate states.
The method's inputs are the user task $\mathcal T$, the current set of candidate states $\mathcal C$ and a generation mode $g \in \{ \texttt{ensemble}, \texttt{independent}, \texttt{conditional} \}$, two of which are inspired from the literature and one is a novel approach. 
Other generation modes certainly exist (e.g., use of different inference configurations; yet results suggest it may be suboptimal \citep{lee2025generatingdiversehypothesesinductive}). Our goal is to empirically evaluate these three distinct design paradigms to establish a foundational understanding of the decoupled exploration-selection framework. 

\textbf{Ensemble generation}: Samples $N$ candidates from a set of $E$ distinct prompts (parameter set by user), instead of relying only on one single prompt. Each prompt is used with probability $p=[p_1,\dots,p_E]$ \footnote{In expectation prompt $i \in \{1, \dots, E\}$ is used $N p_i$ times, with $p_i$ being the probability of using prompt $i$.}. This mode is embarrassingly parallel. Different prompts may access complementary subspaces of the LM's capabilities. This is further motivated by prior research on ensemble learning which improves generalization \citep{dietterich2000ensemble,freund1997decision}. To our knowledge, this ensembling generation mode is novel in test-time scaling of multi-turn agents. Broader impacts include the design of prompt ensembles (chain of thought, few shot, persona-based prompting \citep{olea2024evaluating}), and the learning of optimal $p$ distributions.

We also consider the following baselines. \textbf{Base generation}: Uses the same prompt and inference configuration $N$ times to populate $\mathcal C$. This strategy samples independently from the same LM's stochastic distribution conditioned by the prompt and its inference configuration. This is a baseline strategy previously used in test-time scaling \citep{zhang2025optimizingsequentialmultisteptasks, brown2024largelanguagemonkeysscaling}. Independent generation is embarrassingly parallel and has no sequential dependencies. \textbf{Conditional generation} uses $\mathcal C$ from prior calls to obtain a new candidate $c$ on the current call to \textsc{Generate-Candidate-State}. The prior candidates in $\mathcal C$ are used in the prompt instructions to elicit diversity in the upcoming candidates. Conditional generation is a form of adaptive sampling where each new sample is informed by the previous ones. This generation mode is investigated in prior works \cite{zhang2025optimizingsequentialmultisteptasks,sahu2025insightbenchevaluatingbusinessanalytics}. It is not possible to parallelize in this mode.

\subsection{Selecting states for concurrent execution}

The method \textsc{Select-States} in Algorithm \ref{alg:framework} outputs a set $\mathcal S$ ($|\mathcal S|=M$) of selected states that are to be executed concurrently, receiving the candidate set $\mathcal C$, a budget $M$, the task $\mathcal T$, and a selection mode $v \in \{ \texttt{random}, \texttt{entropy}, \texttt{clustering}, \texttt{alignment} \}$. 
For \textbf{Random selection}, $M$ states are sampled uniformly at random from $\mathcal C$. 
For \textbf{Entropy selection}, each candidate is projected into an embedding space where a greedy maximin algorithm maximizes the selection's entropy using cosine distance \citep{nemhauser1978analysis}. 
Under \textbf{Clustering selection}, candidates are embedded and grouped into $M$ clusters via $k$-means \citep{macqueen1967}, selecting the single candidate closest to each centroid. 
Finally, \textbf{Alignment selection} leverages an LLM judge to analyze all $N$ candidates in $\mathcal C$ and return the $M$ indices most aligned with the task, though high values of $N$ scale up inference costs and risk context window cognitive saturation.

\subsection{Aggregating diverse execution results}

$\textsc{Aggregate}$ in Algorithm \ref{alg:framework} takes as input all the execution results, as well as an aggregation mode $a \in \{ \texttt{self}, \texttt{leader} \}$, and it returns the final response. We investigate whether the capacity of the language model aggregator influences final performance. \textbf{Self-aggregation} employs the same LM capacity across all agent nodes. \textbf{Leader aggregation} selectively upgrades the aggregator node to a more advanced LM. We use identical agent execution logs; the only difference is the model used at the aggregator node. This ensures that all upstream agent states are identical, isolating specifically the effect of the aggregator node.

\section{Results}
\label{sec:results}

We evaluate one agent jointly on open- and closed-ended tasks. Infi-DA-Bench \citep{hu2024infiagent} contributes $257$ closed-ended tasks with verifiable answers, scored by its official rule-based function. InsightBench \citep{sahu2025insightbenchevaluatingbusinessanalytics} contributes $100$ open-ended tasks; performance is the proportion of its $428$ ground-truth insights retrieved, scored by an LLM judge (Claude Opus 4.6). Details are in Appendix~\ref{app:scoring}. We discuss limitations in Section \ref{sec:discussion}.

\textbf{Baselines.} \texttt{DataWise} \citep{you2025datawiseagentnotebookcentricllmagent}, state of the art on Infi-DA-Bench, and \texttt{Agent-Poirot} \citep{sahu2025insightbenchevaluatingbusinessanalytics}, state of the art on InsightBench, both run with their official hyper-parameters. All methods share one inference configuration and three random seeds. Within the decoupled exploration-selection framework, base generation with random selection constitutes the baseline.

\textbf{Models.} Two model suites spanning a proprietary and an open-weights deployment: Claude 4.5 Haiku (aggregator: Claude 4.5 Sonnet) and DeepSeek-V3.2 (aggregator: DeepSeek-R1). We deliberately target the cost-efficient tier of each model family rather than frontier models: enterprise analytics workloads run at volumes where per-task inference cost dominates deployment decisions, and our question is precisely whether test-time scaling lets such models reach specialist-level accuracy.

\definecolor{sigstrong}{HTML}{C6DBEF}
\definecolor{signominal}{HTML}{EAF1F8}
\begin{table}[htb]
\centering
\caption{Main effects, as paired contrasts against the \texttt{Base}+\texttt{Random} default, pooled over both models, both budgets and three seeds. \emph{Agg.}: aggregation arm (\emph{both} pools the two). $\Delta$ in accuracy points, benchmarks weighted equally; CIs and one-sided $p$ from a stratified bootstrap over question clusters. \emph{Wins}: settings with a positive effect. Shading: $p<0.01$ (\colorbox{sigstrong}{darker}), $p<0.05$ (\colorbox{signominal}{lighter}).}
\label{tab:summary_main_effects}
\scriptsize
\renewcommand{\arraystretch}{0.88}
\setlength{\tabcolsep}{4pt}
\begin{tabular}{llrrrcrr}
\toprule
& & \multicolumn{2}{c}{\textbf{$\Delta$ per bench.}} & \multicolumn{4}{c}{\textbf{Overall}} \\
\cmidrule(lr){3-4} \cmidrule(lr){5-8}
\textbf{Contrast} & \textbf{Agg.} & \textbf{Infi} & \textbf{Ins.} & $\mathbf{\Delta}$ & \textbf{95\% CI} & \textbf{Wins} & $\mathbf{p}$ \\
\midrule
\multicolumn{8}{l}{\textit{Generation strategy (vs \texttt{Base})}} \\
\quad Conditional & self & +0.54 & +0.18 & +0.36 & [-0.27, 1.00] & 5/8 & 0.141 \\
\quad  & leader & +0.32 & -0.31 & +0.01 & [-0.61, 0.63] & 3/8 & 0.482 \\
\quad  & both & +0.43 & -0.06 & +0.19 & [-0.26, 0.62] & 8/16 & 0.207 \\
\cmidrule(lr){1-8}
\rowcolor{sigstrong}
\quad Ensemble & self & +6.51 & +0.56 & +3.53 & [2.80, 4.26] & 6/8 & $<$0.001 \\
\rowcolor{sigstrong}
\quad  & leader & +6.36 & +0.02 & +3.19 & [2.50, 3.85] & 6/8 & $<$0.001 \\
\rowcolor{sigstrong}
\quad  & both & +6.44 & +0.29 & +3.36 & [2.86, 3.85] & 12/16 & $<$0.001 \\
\midrule
\multicolumn{8}{l}{\textit{Selection strategy (vs \texttt{Random})}} \\
\quad Clustering & self & +0.12 & +0.44 & +0.28 & [-0.41, 0.94] & 4/8 & 0.198 \\
\quad  & leader & +0.63 & -0.58 & +0.02 & [-0.64, 0.68] & 4/8 & 0.462 \\
\quad  & both & +0.37 & -0.07 & +0.15 & [-0.33, 0.62] & 8/16 & 0.257 \\
\cmidrule(lr){1-8}
\quad Goal-Align & self & +0.13 & +0.58 & +0.35 & [-0.39, 1.06] & 5/8 & 0.169 \\
\quad  & leader & +0.43 & -0.66 & -0.11 & [-0.78, 0.53] & 4/8 & 0.640 \\
\quad  & both & +0.28 & -0.04 & +0.12 & [-0.37, 0.63] & 9/16 & 0.316 \\
\cmidrule(lr){1-8}
\quad MaxEnt & self & +0.23 & +0.83 & +0.53 & [-0.20, 1.26] & 6/8 & 0.076 \\
\quad  & leader & +0.40 & +0.29 & +0.35 & [-0.33, 1.01] & 5/8 & 0.150 \\
\rowcolor{signominal}
\quad  & both & +0.31 & +0.56 & +0.44 & [-0.05, 0.93] & 11/16 & 0.040 \\
\midrule
\multicolumn{8}{l}{\textit{Aggregation (leader-agg.\ vs self-agg.)}} \\
\rowcolor{signominal}
\quad All 12 configurations & -- & +0.69 & +0.24 & +0.46 & [0.04, 0.89] & 4/8 & 0.018 \\
\bottomrule
\end{tabular}
\end{table}

\subsection{Analysis of the main effects of the decoupled exploration-selection framework}

Table~\ref{tab:summary_main_effects} isolates each design axis as a paired contrast against the \texttt{Base}+\texttt{Random} default, measured independently in both aggregation arms and pooled. Per-cell results for all $96$ sweep cells ($12$ configurations $\times$ $2$ budgets $\times$ $2$ models $\times$ $2$ benchmarks), including per-question input/output token usage for every configuration, are in Appendix Tables~\ref{tab:exhaustive_infidabench_haiku45}--\ref{tab:exhaustive_insightbench_deepseekv32}.

\finding{Ensemble has a robust effect on performance, and it pays on closed-ended tasks.} Ensemble generation improves on base generation by ${+}3.36$pp pooled ($95\%$ CI $[2.86, 3.85]$), and the effect replicates in each aggregation arm independently (${+}3.53$ self, ${+}3.19$ leader). Its decomposition is the paper's central asymmetry: ${+}6.44$pp on Infi-DA-Bench against ${+}0.29$pp on InsightBench. Conditional generation, despite its sequential cost, is equivalent to base (${+}0.19$pp, $p=0.207$).

\finding{Upgrading the aggregator improves performance.} Holding the candidate plans fixed and swapping only the aggregating model, leader aggregation gains ${+}0.46$pp over self-aggregation ($95\%$ CI $[0.04, 0.89]$, $p=0.018$). Although an order of magnitude below the generation effect, this points to the potential of resource-aware agentic designs that mix LLM capacities across nodes, reserving the stronger (more expensive) model for the single node where it measurably pays.

\finding{No selection strategy separates from uniform random choice.} The largest selection effect is MaxEnt at ${+}0.44$pp ($95\%$ CI $[-0.05, 0.93]$), and is significant ($p=0.040$). However, the magnitude of the effect is small, and Clustering and Goal-Align sit within $\pm 0.2$pp of random. Since random selection alone adds zero cost --- the alternatives spend $N$ embedding calls or a full LLM pass over every candidate --- it is the deployment default. Plan selection is thus the one axis where withholding compute costs nothing.

\subsection{Analysis of absolute performance and baseline comparison}

\finding{The best configuration.} Ranking all $48$ configurations (Table~\ref{tab:top_configs}; full ranking in Appendix~\ref{app:refconfig}), \textbf{\texttt{Ensemble}+\texttt{Random} at $M{=}5$ with leader aggregation} ranks first and attains the highest overall score in $57.7\%$ of paired bootstrap resamples --- far above the $2.1\%$ chance level, but short of an identified winner: five other configurations, all \texttt{Ensemble} at $M{=}5$, are statistically indistinguishable from it ($p>0.05$). Every one of them pays for a selector that Table~\ref{tab:summary_main_effects} shows to have marginal returns, so we adopt the cheapest member of the tied group as the reference configuration$^\dagger$: it requires no tuning and no selection overhead. The gap to the default is large: the \texttt{Base}+\texttt{Random} single-prompt baseline at the same budget ranks $11$th and never wins a resample.

\finding{One \texttt{CIPHER} configuration matches both purpose-built specialist baselines.} Under matched base models (Table~\ref{tab:results}), the reference configuration reaches \texttt{DataWise}, the closed-ended state of the art, within $1.2$pp on both suites, while lifting the single-plan agent from $76.5$ to $82.5\%$ two-suite mean; and it brackets \texttt{Agent-Poirot}, the open-ended state of the art, at ${+}1.1$pp on Haiku 4.5 and ${-}1.2$pp on DeepSeek-V3.2. Neither specialist transfers: \texttt{DataWise} has no mechanism for open-ended insight discovery and \texttt{Agent-Poirot} none for verifiable extraction, whereas the same \texttt{CIPHER} configuration serves both workloads at specialist-level accuracy.

\begin{table}[htb]
\centering
\caption{Top-3 of the $48$-point configuration space (generation, selection, $M$, aggregation arm) against the \texttt{Base}+\texttt{Random} baseline. Accuracy averages both base models, three seeds and both benchmarks with equal weight. \emph{P(best)} is the share of $10000$ paired bootstrap resamples (question clusters, shared across rows) in which a row scores highest. Full ranking: Table~\ref{tab:config_space}, Appendix.}
\label{tab:top_configs}
\scriptsize
\renewcommand{\arraystretch}{0.88}
\setlength{\tabcolsep}{4pt}
\begin{tabular}{rllclrrrr}
\toprule
& \multicolumn{4}{l}{\textbf{Configuration}} & \multicolumn{3}{c}{\textbf{Accuracy (\%)}} & \\
\cmidrule(lr){2-5} \cmidrule(lr){6-8}
\textbf{\#} & \textbf{Gen.} & \textbf{Sel.} & $\mathbf{M}$ & \textbf{Agg.} & \textbf{Infi} & \textbf{Ins.} & \textbf{Overall} & \textbf{P(best)} \\
\midrule
\rowcolor{goldrow}
1 & Ensemble & Random & 5 & leader & 82.49 & 51.87 & \textbf{67.18} & 57.7\% \\
\rowcolor{silverrow}
2 & Ensemble & Goal-Align & 5 & leader & 82.36 & 51.21 & 66.78 & 20.8\% \\
\rowcolor{bronzerow}
3 & Ensemble & Goal-Align & 5 & self & 82.49 & 50.55 & 66.52 & 9.6\% \\
11 & Base & Random & 5 & leader & 76.52 & 51.25 & 63.89 & $<$0.1\% \\
\bottomrule
\end{tabular}
\end{table}

\begin{table}[htb]
\centering
\caption{\texttt{CIPHER} against purpose-built specialist agents, over three seeds. $^\dagger$: the reference configuration (Ensemble, random selection, leader aggregation, $M{=}5$; App.~\ref{app:refconfig}).}
\label{tab:results}
\scriptsize
\renewcommand{\arraystretch}{0.88}
\begin{tabular}{@{}llc@{\hspace{10pt}}llc@{}}
\toprule
\multicolumn{3}{c}{\textbf{Infi-DA-Bench (closed-ended)}} & \multicolumn{3}{c}{\textbf{InsightBench (open-ended)}} \\
\cmidrule(r){1-3} \cmidrule(l){4-6}
\textbf{Model} & \textbf{Method} & \textbf{Acc.(\%)} & \textbf{Model} & \textbf{Method} & \textbf{Acc.(\%)} \\
\midrule
\multirow{2}{*}{Haiku-4.5} & DataWise & 84.18 {\tiny$\pm$0.59} & \multirow{2}{*}{Haiku-4.5} & Agent-Poirot & 54.05 {\tiny$\pm$0.97} \\
 & CIPHER$^\dagger$(10,5) & 83.40 {\tiny$\pm$0.45} &  & CIPHER$^\dagger$(10,5) & 55.14 {\tiny$\pm$2.00} \\
\cmidrule(r){1-3} \cmidrule(l){4-6}
\multirow{2}{*}{DeepSeek-V3.2} & DataWise & 82.75 {\tiny$\pm$1.12} & \multirow{2}{*}{DeepSeek-V3.2} & Agent-Poirot & 49.84 {\tiny$\pm$1.89} \\
 & CIPHER$^\dagger$(10,5) & 81.58 {\tiny$\pm$0.59} &  & CIPHER$^\dagger$(10,5) & 48.60 {\tiny$\pm$0.47} \\
\bottomrule
\end{tabular}
\end{table}

\bibliographystyle{plainnat}
\bibliography{references}

\appendix

\section{Related Works}
\label{sec:related_works}

Our design connects to a broader test-time scaling literature. \citet{wang2023selfconsistency} established parallel sampling as a reliable inference-time technique on reasoning benchmarks. \citet{yao2023tree} generalize this into structured search over reasoning trees, and \citet{du2024debate} propose parallel agent debate for open-ended reasoning. Sequential test-time refinement (\citealp{madaan2023selfrefine}, \citealp{shinn2023reflexion}) offers a complementary alternative that trades parallel breadth for iterative depth. Recent work by \citet{snell2025scaling} characterizes optimal test-time compute allocation, providing theoretical context for our empirical study of the generation--selection trade-off.

\citep{brown2024largelanguagemonkeysscaling} shows that multiple executions of the same prompt in single-turn LM applications improves performance at test-time on closed-ended tasks. \citep{anonymous2025generalized} proposes a system to leverage parallel execution of LLM prompts with inter-communication to improve performance and share information across threads.
\cite{wang2025thinkdeepthinkfast} compare the returns of parallelization versus reasoning on large language models and finds that reasoning tends to outperform parallelization. However, \cite{liu20251bllmsurpass405b} show that test-time scaling with the right generation, selection and aggregation strategy can outperform larger LM models, highlighting that parallelization is an avenue for cost efficient designs.
\cite{inoue2025widerdeeperscalingllm} highlight that adaptive strategies that emulate diverse solution paths improve performance, and propose a method based on exploration-exploitation.

Although test-time scaling properties have been investigated in the realm of single-turn LM applications, its impact on agentic designs remains an emerging avenue of research.
\cite{qin2025flashsearcherfasteffectiveweb} propose Flash-Searcher, showing how parallelization can deliver latency speed-ups when sectioning problems into independent parts.
\citep{zhu2025scalingtesttimecomputellm} operate concurrent executions using diverse LLMs.
\cite{zhang2025optimizingsequentialmultisteptasks} proposes a seminal framework for test-time scaling but there is no clear distinction between generation and selection. Furthermore, many questions remain open regarding the number of agents to parallelize, and the strategy to aggregate results. The approach SPIO \citep{seo2026spioensembleselectivestrategies} implements test-time scaling at the node level where each node is executed multiple times within an agentic workflow.

Prior works \citep{anonymous2025think} on single-turn LM test-time scaling propose to execute multiple concurrent executions and then use softmax averaging over the distribution of tokens to aggregate and obtain a final response. The limitation of this approach is that it assumes access to the logits of the LM, which is often not the case in real-world applications where LMs are accessed through APIs.

\section{Evaluation setup}
\label{app:scoring}

We evaluate one agent jointly on open- and closed-ended tasks, which to our knowledge has not been done before. We conduct experiments on a total of $357$ tasks on Infi-DA-Bench \citep{hu2024infiagent} and InsightBench \citep{sahu2025insightbenchevaluatingbusinessanalytics}. 
InsightBench \citep{sahu2025insightbenchevaluatingbusinessanalytics} evaluates performance on open-ended ambiguous tasks (e.g., hypothesis validation through data analysis). 
In contrast, Infi-DA-Bench \citep{hu2024infiagent} evaluates performance on closed-ended tasks (e.g., exploratory data analysis and modeling). 
Therefore, Infi-DA-Bench corresponds to only a subspace of the skills required to solve InsightBench. Examples of tasks from both benchmarks are available in Table \ref{tab:task-examples}.

\begin{table}[htb]
\caption{Examples of closed-ended and open-ended tasks used in our evaluation setup.}
\label{tab:task-examples}
\centering
\small
\begin{tabular}{@{}p{\linewidth}@{}}
\toprule
\textbf{Closed-ended (Infi-DA-Bench)} \\
Identify countries where \texttt{gdpPercap\_1982} values are outliers using the IQR method ($1.5 \times$ IQR threshold). Return the list in format: \texttt{@outlier\_countries[list]}. \\
\midrule
\textbf{Open-ended (InsightBench)} \\
Analyze warranty period distribution by model category, cost distribution by model category, and the correlation between cost and warranty period for computer assets. Provide 5 data-driven insights. \\
\bottomrule
\end{tabular}
\end{table}

\subsection{Open-ended tasks (InsightBench)}

InsightBench \citep{sahu2025insightbenchevaluatingbusinessanalytics} includes a set of $T=100$ tasks.
Each task comes with a dataset and a set of ground truth actionable insights.
The goal of the agent on each task is to retrieve accurately the ground truth actionable insights in the dataset.
There is a total of $428$ ground truth insights to be retrieved across all the questions. 

\textbf{Task setup} To help the agent retrieve the ground truth insights, we provide the agent with a set of exploratory hypotheses (formulated as open-ended questions). 
This is akin to a data scientist having formulated a set of exploratory hypotheses and using the agent to conduct the analysis. 
For reproducibility, the exploratory hypotheses we use are those available in the meta-data of each task in the benchmark.
Another possible setup consists in evaluating the agent to retrieve the ground truth insights without the guidance of the exploratory hypotheses. 
As evidenced by our empirical results, both setups are challenging, with the full autonomy setup being harder.

\textbf{Performance metric}
The performance is defined as the proportion of insights correctly retrieved by the agent across the $T$ questions and $428$ insights:
$ \text{Accuracy} =  \frac{\sum_{i=1}^{T} \text{Insights Retrieved}_i}{\text{Total Insights}} $.
An increase of $1$pp corresponds to retrieving approximately $4.3$ insights. While semantic similarity metrics have been proposed for this benchmark \citep{sahu2025insightbenchevaluatingbusinessanalytics}, they are difficult to interpret and do not directly reflect the agent's operational ability to retrieve insights, unlike the accuracy metric we adopt.

\textbf{Correctness evaluation}
A LLM-as-a-judge \citep{Zheng2023llmjudge} evaluates whether the agent's generated insights match the ground truth. We use Claude Opus 4.6 as the judge, emitting one boolean verdict per ground-truth insight. The judge is held fixed across all methods and both aggregation arms, so relative comparisons are unbiased with respect to the choice of judge.

\textbf{State-of-the-art baseline} The current state-of-the art on InsightBench is the open-source strategy \texttt{Agent-Poirot} \citep{sahu2025insightbenchevaluatingbusinessanalytics}. We evaluate \texttt{Agent-Poirot} with the official repository hyper-parameters. 

\subsection{Close-ended tasks (Infi-DA-Bench)}

Infi-DA-Bench benchmark includes a total of $T=257$ tasks. Each question is composed of a dataset and a closed-ended question (e.g. descriptive statistics, modeling). The goal of the agent is to respond correctly to the question.

\textbf{Performance metric} We report the performance as the accuracy over the set of $T$ tasks, measured as 
$ \text{Accuracy} =  \frac{ 1}{T} \sum_{i=1}^{T} [ \text{Correct Response}_i ] $.
An increase of $1$pp corresponds to the correct resolution of approximately $2.57$ tasks. 

\textbf{Correctness evaluation} We use the official rule-based evaluation function to evaluate the responses \citep{hu2024infiagent}. The function is strict: small numerical deviation from the ground truth (up to $1e-6$), partially true responses, or valid responses with wrong formatting are all marked as incorrect. Consequently, performance improvements under this strict evaluation are meaningful and significant.

\textbf{State-of-the-art baseline} The current state of the art is \texttt{Data-Wise} agent \cite{you2025datawiseagentnotebookcentricllmagent}. Other agents exist but have been outperformed by \texttt{Data-Wise} agent \citep{wu2024autogen,hong2024datainterpreterllmagent}. We consider \texttt{Data-Wise} agent as a baseline, and we use the official hyper-parameter configuration reported in the open-source repository.

\section{Reproducibility details}

The main algorithm is illustrated in Algorithm \ref{alg:framework}. 
We highlight in \colorbox{green!20}{green} the main technical novelty compared to existing frameworks~\citep{zhang2025optimizingsequentialmultisteptasks}.

\begin{algorithm}[t]
\caption{Decoupled Exploration-Selection (DES) framework}
\label{alg:framework}
\begin{algorithmic}[1]
\Require Task $\mathcal{T}$, budget $N$ and $M$, generation mode $g$, selection mode $v$, aggregation mode $a$
\Ensure Final state $s^*$
\State \textbf{Generate candidate states:}
\State $\mathcal{C} \leftarrow \{\}$
\For{$n \in \{1, 2, \dots, N\}$}
    \State \colorbox{green!20}{$c \leftarrow \textsc{Generate-Candidates}(\mathcal{T}, \mathcal{C}, g)$}
    \State $\mathcal{C} \leftarrow \mathcal{C} \cup \{c\}$
\EndFor
\State \textbf{Select initial states:}
\State \colorbox{green!20}{$\mathcal{S} \leftarrow \textsc{Select-States}(\mathcal{T}, \mathcal{C}, M, v)$}
\State \textbf{Execute initial states (conceptually in parallel):}
\State $\mathcal{P} \leftarrow \{\}$
\For{each $s \in \mathcal{S}$}
    \State $\tau \leftarrow \textsc{Execute-Initial-State}(\mathcal{T}, s)$
    \State $\mathcal{P} \leftarrow \mathcal{P} \cup \{\tau\}$
\EndFor
\State \textbf{Aggregate results:}
\State \colorbox{green!20}{$s^* \leftarrow \textsc{Aggregate}(\mathcal{P}, a)$}
\State \Return $s^*$
\end{algorithmic}
\end{algorithm}

\textbf{Trade-off related to cost efficiency}
$\texttt{CIPHER}$ is instructed to reason exclusively on the text modality. Although the generated code can produce visual artifacts (e.g., charts, plots), the downstream agents only consume the corresponding textual logs (e.g., summary statistics, model metrics, tables) printed during code execution. This design choice is motivated by two key factors: i) high performance on images necessitates multimodal reasoning LMs which are more costly, ii) image tokenization costs are orders of magnitude greater than processing equivalent information through text.

\textbf{Aggregator node} We adopt an open-source prompt previously proposed to aggregate outcomes from concurrent executions \citep{zhang2025optimizingsequentialmultisteptasks}. 

\subsection{Hyper-parameters}

All agents (baselines included) share one inference configuration: temperature
$0.5$, top-p $0.99$, top-k $500$, chosen via a sensitivity study. The two base
suites reported are Claude 4.5 Haiku and DeepSeek-V3.2, each paired with a
stronger leader aggregator from the same family (Claude 4.5 Sonnet, DeepSeek-R1).
Baselines use their official repository hyper-parameters. Each
configuration is evaluated on three random seeds. The judge used to evaluate agent findings on InsightBench is Opus 4.6.

\subsection{Implementation details on the generation strategies}

\textbf{Ensemble strategy} The choice of the optimal $E$ parameter is linked to the parameter $M$. For a task, if $M$ states are executed, the set of prompts $E$ is more likely to be fully covered when $E<M$. Since in our experiments we focus on $M=3$ and $M=5$ regimes, we set $E=2$ and $p=[0.5,0.5]$. The two prompts are: prompt 1) is the base prompt used in CIPHER (1,1) the base agent, 2) the second prompt is a paraphrase obtained from a LLM. The key difference is that the paraphrase in prompt no. 2 imposes a fixed structure to the plan generated, while prompt no. 1 does not. 

\subsection{Implementation details on the selection strategies}

All embedding-based methods use Amazon Titan Text Embeddings v2 (\texttt{amazon.titan-embed-text-v2:0}), which produce unit-normalized vectors; no additional normalization is applied. The dimensions is set to $1,024$ as embedding size. To our knowledge, projecting candidate spaces into a latent space for subset selection is a novel algorithmic approach for test-time scaling AI agents. Designing embedding spaces that further capture semantic properties of candidates is a promising avenue for future work.

\textbf{Random.} $M$ plans are sampled uniformly at random without replacement.

\textbf{MaxEnt (Greedy Maximin).} We compute pairwise cosine distances between all $N$ plan embeddings. Selection proceeds greedily: the two most distant plans are chosen first, then each subsequent plan is the one maximizing its minimum cosine distance to any already-selected plan. This greedy maximin procedure approximates maximum entropy over the selected set.

\textbf{Clustering.} We apply $k$-means with $k{=}M$ clusters over the plan embeddings (scikit-learn implementation, $k$-means++ initialization, $n\_init{=}10$ restarts, seeded with the experiment random seed). Within each cluster, the plan closest to the centroid in Euclidean distance is selected as the representative.

\textbf{Goal Align (LLM Judge).} An LLM judge (Claude 3.5 Sonnet, temperature 0.5) receives the original question and all $N$ plan texts, and is prompted to select the $M$ most promising plans based on three criteria: goal alignment, feasibility, and diversity of perspectives. The judge outputs a JSON array of $M$ plan indices. The full prompt context comprises the question and all plans without truncation; the response is limited to 200 tokens. If parsing fails, the method falls back to random selection.

\section{Additional results}
\label{app:additional}

\begin{figure}[htb]
    \centering
    \includegraphics[width=0.99\linewidth]{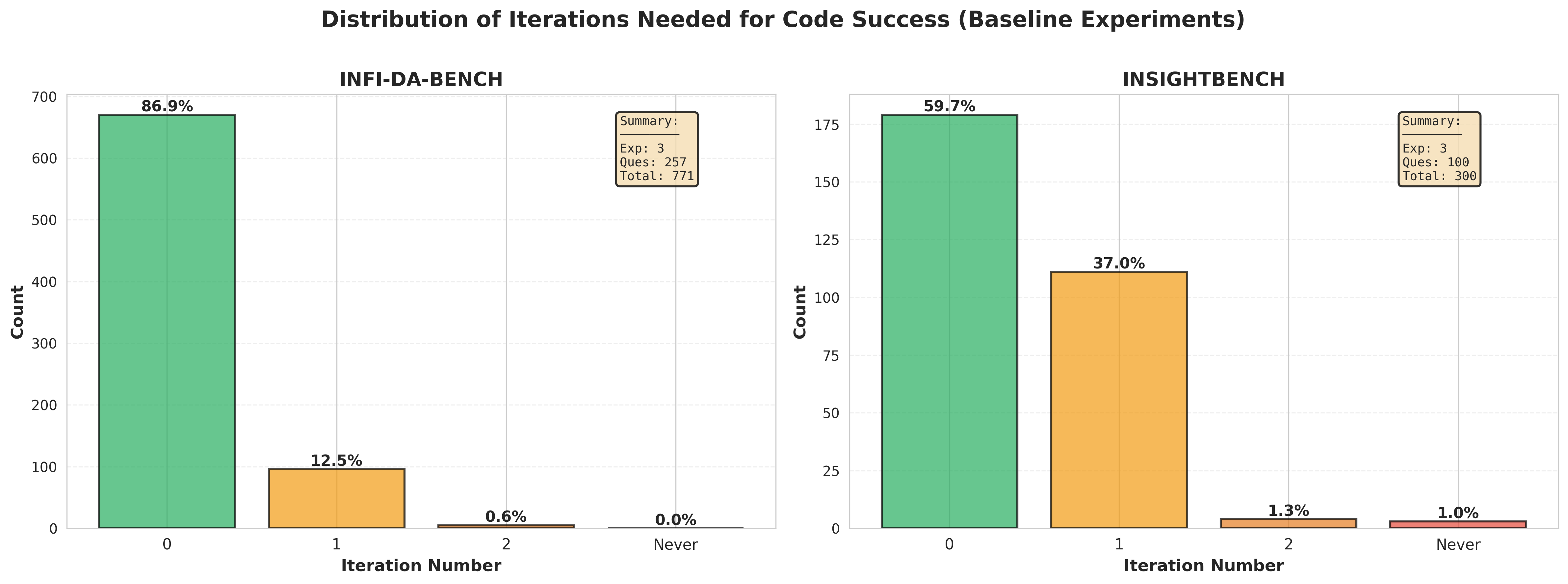}
    \caption{Distribution of iterations needed to reach executable code across \texttt{CIPHER}(1,1) baseline experiments on three random seeds for Infi-DA-Bench (left) and InsightBench (right).}
    \label{fig:coding_loop}
\end{figure}

\textbf{Ablation on CIPHER coding loop design}
Within each plan step, if code execution fails, the router sends the failed cell to the \texttt{editor} for a fix; the retry loop \texttt{coder} $\rightarrow$ \texttt{executor} $\rightarrow$ \texttt{analyzer} $\rightarrow$ \texttt{editor} $\rightarrow$ \texttt{executor} can repeat up to three times before the step is routed to the \texttt{finalizer}. We set \texttt{max\_iterations}$=3$ based on the observed distribution of retries under our baseline \texttt{CIPHER}(1,1) configuration (Figure~\ref{fig:coding_loop}): $86.9\%$ of steps on Infi-DA-Bench and $59.7\%$ on InsightBench succeed on the first attempt (iteration $0$), and by the second retry (iteration $2$) the cumulative success rate reaches $\geq99\%$ on both benchmarks. Only $\leq1\%$ of steps ever exhaust the loop without success. A cap of three iterations therefore captures essentially all recoverable failures while bounding the worst-case token cost per step.

\subsection{Result details for Haiku 4.5}

Tables~\ref{tab:exhaustive_infidabench_haiku45} and~\ref{tab:exhaustive_insightbench_haiku45} report every cell of the sweep for the Claude 4.5 Haiku suite: all $12$ generation $\times$ selection configurations at both budgets, with both aggregation arms, plan-diversity entropy, and per-question token usage. These are the per-cell measurements underlying the pooled contrasts of Table~\ref{tab:summary_main_effects}; the entropy columns additionally show \emph{why} the selection axis is null --- MaxEnt visibly raises the entropy of the selected set (e.g., $0.118 \rightarrow 0.182$) without a corresponding accuracy gain, so selected-plan diversity is not the binding constraint.

\begin{table*}[htb]
\centering
\caption{Results on \textbf{InFi-DA-Bench} with Claude 4.5 Haiku ($N{=}10$). We report self-aggregation and leader-aggregation accuracy (mean $\pm$ std), entropy for all plans $\rightarrow$ selected plans, and average token usage per question (in thousands) across three random seeds. Bold marks the best accuracy per column within each generation block. Both aggregation arms use identical candidate plans and differ only in the aggregating model, and on InFi-DA-Bench both arms are scored by the same evaluator, so the two accuracy columns are directly comparable. Entropy is computed from Titan embeddings of the plan texts.}
\label{tab:exhaustive_infidabench_haiku45}
\resizebox{\textwidth}{!}{
\begin{tabular}{ll|cccc|cccc}
\toprule
& & \multicolumn{4}{c|}{\textbf{M = 3}} & \multicolumn{4}{c}{\textbf{M = 5}} \\
\cmidrule(lr){3-6} \cmidrule(lr){7-10}
& & \multicolumn{2}{c}{\textbf{Accuracy (\%)}} & & & \multicolumn{2}{c}{\textbf{Accuracy (\%)}} & & \\
\cmidrule(lr){3-4} \cmidrule(lr){7-8}
\textbf{Generation} & \textbf{Selection} & \textbf{self-agg.} & \textbf{leader-agg.} & \textbf{Entropy} & \textbf{Tok In/Out (K)} & \textbf{self-agg.} & \textbf{leader-agg.} & \textbf{Entropy} & \textbf{Tok In/Out (K)} \\
\midrule
Base & Clustering & 63.16 {\scriptsize $\pm$ 3.31} & 64.98 {\scriptsize $\pm$ 3.18} & 0.118 $\rightarrow$ 0.138 & 156.1/25.4 & \textbf{72.37 {\scriptsize $\pm$ 1.78}} & 75.88 {\scriptsize $\pm$ 1.35} & 0.118 $\rightarrow$ 0.136 & 216.0/37.0 \\
 & Goal Align & 62.78 {\scriptsize $\pm$ 0.22} & 64.72 {\scriptsize $\pm$ 0.98} & 0.117 $\rightarrow$ 0.115 & 157.9/25.6 & 71.60 {\scriptsize $\pm$ 3.57} & 74.45 {\scriptsize $\pm$ 1.84} & 0.118 $\rightarrow$ 0.118 & 214.1/36.8 \\
 & Maxent & \textbf{64.85 {\scriptsize $\pm$ 1.19}} & \textbf{66.54 {\scriptsize $\pm$ 2.06}} & 0.118 $\rightarrow$ 0.182 & 155.1/25.2 & 71.98 {\scriptsize $\pm$ 2.81} & 74.19 {\scriptsize $\pm$ 1.12} & 0.118 $\rightarrow$ 0.148 & 217.6/36.9 \\
 & Random & 63.29 {\scriptsize $\pm$ 2.00} & 64.20 {\scriptsize $\pm$ 2.06} & 0.118 $\rightarrow$ 0.116 & 158.5/25.6 & 72.24 {\scriptsize $\pm$ 1.57} & \textbf{76.65 {\scriptsize $\pm$ 2.70}} & 0.117 $\rightarrow$ 0.117 & 219.5/37.4 \\
\cmidrule(lr){2-10}
\multicolumn{2}{l|}{\textit{Base Average}} & \textit{63.52 {\scriptsize $\pm$ 0.91}} & \textit{65.11 {\scriptsize $\pm$ 1.00}} & \textit{0.118 $\rightarrow$ 0.138} & \textit{156.9/25.5} & \textit{72.05 {\scriptsize $\pm$ 0.34}} & \textit{75.29 {\scriptsize $\pm$ 1.17}} & \textit{0.118 $\rightarrow$ 0.130} & \textit{216.8/37.0} \\
\midrule
Conditional & Clustering & 65.89 {\scriptsize $\pm$ 0.59} & \textbf{67.96 {\scriptsize $\pm$ 0.81}} & 0.119 $\rightarrow$ 0.139 & 157.3/25.3 & \textbf{74.45 {\scriptsize $\pm$ 3.17}} & \textbf{76.78 {\scriptsize $\pm$ 0.59}} & 0.119 $\rightarrow$ 0.137 & 216.5/37.0 \\
 & Goal Align & 64.85 {\scriptsize $\pm$ 0.22} & 64.85 {\scriptsize $\pm$ 1.62} & 0.119 $\rightarrow$ 0.119 & 158.7/25.6 & 71.08 {\scriptsize $\pm$ 0.22} & 75.62 {\scriptsize $\pm$ 0.59} & 0.119 $\rightarrow$ 0.117 & 215.3/36.9 \\
 & Maxent & 63.29 {\scriptsize $\pm$ 4.20} & 66.02 {\scriptsize $\pm$ 3.84} & 0.119 $\rightarrow$ 0.184 & 158.3/25.6 & 73.41 {\scriptsize $\pm$ 1.19} & 75.62 {\scriptsize $\pm$ 1.37} & 0.118 $\rightarrow$ 0.149 & 216.3/37.0 \\
 & Random & \textbf{66.02 {\scriptsize $\pm$ 3.99}} & 66.80 {\scriptsize $\pm$ 2.97} & 0.118 $\rightarrow$ 0.121 & 157.8/25.6 & 73.02 {\scriptsize $\pm$ 3.68} & 75.23 {\scriptsize $\pm$ 3.82} & 0.117 $\rightarrow$ 0.118 & 214.5/36.9 \\
\cmidrule(lr){2-10}
\multicolumn{2}{l|}{\textit{Conditional Average}} & \textit{65.01 {\scriptsize $\pm$ 1.26}} & \textit{66.41 {\scriptsize $\pm$ 1.31}} & \textit{0.119 $\rightarrow$ 0.141} & \textit{158.0/25.5} & \textit{72.99 {\scriptsize $\pm$ 1.41}} & \textit{75.81 {\scriptsize $\pm$ 0.67}} & \textit{0.118 $\rightarrow$ 0.130} & \textit{215.7/37.0} \\
\midrule
Ensemble & Clustering & \textbf{76.91 {\scriptsize $\pm$ 3.31}} & 78.47 {\scriptsize $\pm$ 3.26} & 0.108 $\rightarrow$ 0.133 & 164.9/26.3 & 79.12 {\scriptsize $\pm$ 1.19} & 82.10 {\scriptsize $\pm$ 1.70} & 0.108 $\rightarrow$ 0.128 & 221.6/37.6 \\
 & Goal Align & 76.78 {\scriptsize $\pm$ 3.02} & \textbf{79.25 {\scriptsize $\pm$ 3.99}} & 0.106 $\rightarrow$ 0.102 & 170.8/26.5 & \textbf{84.18 {\scriptsize $\pm$ 1.92}} & \textbf{84.95 {\scriptsize $\pm$ 1.47}} & 0.108 $\rightarrow$ 0.101 & 237.0/39.0 \\
 & Maxent & 75.62 {\scriptsize $\pm$ 2.81} & 77.43 {\scriptsize $\pm$ 1.78} & 0.108 $\rightarrow$ 0.170 & 163.9/26.1 & 79.64 {\scriptsize $\pm$ 2.14} & 82.36 {\scriptsize $\pm$ 0.22} & 0.108 $\rightarrow$ 0.138 & 222.7/37.7 \\
 & Random & 74.32 {\scriptsize $\pm$ 3.75} & 75.23 {\scriptsize $\pm$ 3.26} & 0.108 $\rightarrow$ 0.108 & 165.1/26.2 & 81.71 {\scriptsize $\pm$ 0.39} & 83.40 {\scriptsize $\pm$ 0.45} & 0.108 $\rightarrow$ 0.107 & 223.3/37.8 \\
\cmidrule(lr){2-10}
\multicolumn{2}{l|}{\textit{Ensemble Average}} & \textit{75.91 {\scriptsize $\pm$ 1.21}} & \textit{77.59 {\scriptsize $\pm$ 1.74}} & \textit{0.108 $\rightarrow$ 0.128} & \textit{166.2/26.3} & \textit{81.16 {\scriptsize $\pm$ 2.30}} & \textit{83.20 {\scriptsize $\pm$ 1.29}} & \textit{0.108 $\rightarrow$ 0.119} & \textit{226.2/38.0} \\
\bottomrule
\end{tabular}
}
\end{table*}

\begin{table*}[htb]
\centering
\caption{Results on \textbf{InsightBench} with Claude 4.5 Haiku ($N{=}10$). Accuracy is the proportion of the 428 ground-truth insights retrieved across the 100 questions, so one percentage point corresponds to $\approx$4.3 insights. We report self-aggregation and leader-aggregation accuracy (mean $\pm$ std), entropy for all plans $\rightarrow$ selected plans, and average token usage per question (in thousands) across three random seeds. Bold marks the best accuracy per column within each generation block. Both aggregation arms use identical candidate plans and differ only in the aggregating model, and on InsightBench both arms are scored by the same evaluator, so the two accuracy columns are directly comparable. Entropy is computed from Titan embeddings of the plan texts.}
\label{tab:exhaustive_insightbench_haiku45}
\resizebox{\textwidth}{!}{
\begin{tabular}{ll|cccc|cccc}
\toprule
& & \multicolumn{4}{c|}{\textbf{M = 3}} & \multicolumn{4}{c}{\textbf{M = 5}} \\
\cmidrule(lr){3-6} \cmidrule(lr){7-10}
& & \multicolumn{2}{c}{\textbf{Accuracy (\%)}} & & & \multicolumn{2}{c}{\textbf{Accuracy (\%)}} & & \\
\cmidrule(lr){3-4} \cmidrule(lr){7-8}
\textbf{Generation} & \textbf{Selection} & \textbf{self-agg.} & \textbf{leader-agg.} & \textbf{Entropy} & \textbf{Tok In/Out (K)} & \textbf{self-agg.} & \textbf{leader-agg.} & \textbf{Entropy} & \textbf{Tok In/Out (K)} \\
\midrule
Base & Clustering & \textbf{54.13 {\scriptsize $\pm$ 0.49}} & \textbf{54.28 {\scriptsize $\pm$ 0.36}} & 0.116 $\rightarrow$ 0.130 & 313.3/54.7 & 51.40 {\scriptsize $\pm$ 2.63} & 52.57 {\scriptsize $\pm$ 2.43} & 0.117 $\rightarrow$ 0.130 & 460.1/81.3 \\
 & Goal Align & 53.66 {\scriptsize $\pm$ 3.17} & 54.05 {\scriptsize $\pm$ 1.78} & 0.114 $\rightarrow$ 0.103 & 325.5/55.3 & 52.41 {\scriptsize $\pm$ 2.86} & 52.10 {\scriptsize $\pm$ 1.87} & 0.117 $\rightarrow$ 0.111 & 450.7/79.6 \\
 & Maxent & 53.35 {\scriptsize $\pm$ 4.40} & 53.58 {\scriptsize $\pm$ 1.66} & 0.118 $\rightarrow$ 0.180 & 325.6/55.9 & \textbf{54.05 {\scriptsize $\pm$ 2.01}} & \textbf{56.15 {\scriptsize $\pm$ 0.75}} & 0.116 $\rightarrow$ 0.145 & 437.3/79.9 \\
 & Random & 51.71 {\scriptsize $\pm$ 1.15} & 51.25 {\scriptsize $\pm$ 2.70} & 0.116 $\rightarrow$ 0.117 & 311.7/54.0 & 52.96 {\scriptsize $\pm$ 2.58} & 54.91 {\scriptsize $\pm$ 2.30} & 0.117 $\rightarrow$ 0.118 & 455.5/80.7 \\
\cmidrule(lr){2-10}
\multicolumn{2}{l|}{\textit{Base Average}} & \textit{53.21 {\scriptsize $\pm$ 1.05}} & \textit{53.29 {\scriptsize $\pm$ 1.39}} & \textit{0.116 $\rightarrow$ 0.132} & \textit{319.0/55.0} & \textit{52.71 {\scriptsize $\pm$ 1.10}} & \textit{53.93 {\scriptsize $\pm$ 1.92}} & \textit{0.117 $\rightarrow$ 0.126} & \textit{450.9/80.4} \\
\midrule
Conditional & Clustering & \textbf{53.66 {\scriptsize $\pm$ 0.36}} & \textbf{53.66 {\scriptsize $\pm$ 0.59}} & 0.115 $\rightarrow$ 0.131 & 317.0/54.6 & 53.19 {\scriptsize $\pm$ 3.40} & 51.95 {\scriptsize $\pm$ 1.05} & 0.115 $\rightarrow$ 0.129 & 468.2/78.3 \\
 & Goal Align & 53.12 {\scriptsize $\pm$ 0.49} & 51.01 {\scriptsize $\pm$ 1.96} & 0.118 $\rightarrow$ 0.111 & 315.2/53.1 & \textbf{54.13 {\scriptsize $\pm$ 1.18}} & 53.66 {\scriptsize $\pm$ 2.96} & 0.117 $\rightarrow$ 0.111 & 472.4/81.5 \\
 & Maxent & 52.80 {\scriptsize $\pm$ 4.05} & 52.02 {\scriptsize $\pm$ 2.36} & 0.116 $\rightarrow$ 0.177 & 310.2/53.2 & 52.88 {\scriptsize $\pm$ 0.88} & \textbf{55.84 {\scriptsize $\pm$ 0.47}} & 0.117 $\rightarrow$ 0.146 & 468.5/80.9 \\
 & Random & 50.86 {\scriptsize $\pm$ 3.65} & 53.04 {\scriptsize $\pm$ 1.53} & 0.116 $\rightarrow$ 0.118 & 311.8/54.0 & 53.27 {\scriptsize $\pm$ 2.69} & 55.69 {\scriptsize $\pm$ 1.72} & 0.115 $\rightarrow$ 0.116 & 464.7/81.7 \\
\cmidrule(lr){2-10}
\multicolumn{2}{l|}{\textit{Conditional Average}} & \textit{52.61 {\scriptsize $\pm$ 1.22}} & \textit{52.43 {\scriptsize $\pm$ 1.16}} & \textit{0.116 $\rightarrow$ 0.134} & \textit{313.5/53.7} & \textit{53.37 {\scriptsize $\pm$ 0.53}} & \textit{54.28 {\scriptsize $\pm$ 1.85}} & \textit{0.116 $\rightarrow$ 0.126} & \textit{468.5/80.6} \\
\midrule
Ensemble & Clustering & 51.87 {\scriptsize $\pm$ 1.07} & 50.93 {\scriptsize $\pm$ 2.65} & 0.128 $\rightarrow$ 0.148 & 287.4/51.9 & 53.97 {\scriptsize $\pm$ 3.67} & 54.83 {\scriptsize $\pm$ 1.35} & 0.129 $\rightarrow$ 0.145 & 489.8/88.0 \\
 & Goal Align & 50.86 {\scriptsize $\pm$ 1.08} & 51.17 {\scriptsize $\pm$ 1.46} & 0.131 $\rightarrow$ 0.088 & 325.3/55.2 & 54.21 {\scriptsize $\pm$ 1.91} & 54.28 {\scriptsize $\pm$ 2.83} & 0.131 $\rightarrow$ 0.099 & 538.8/90.3 \\
 & Maxent & \textbf{54.13 {\scriptsize $\pm$ 2.50}} & \textbf{54.83 {\scriptsize $\pm$ 2.29}} & 0.129 $\rightarrow$ 0.195 & 297.0/54.2 & \textbf{54.28 {\scriptsize $\pm$ 1.18}} & \textbf{55.53 {\scriptsize $\pm$ 2.74}} & 0.130 $\rightarrow$ 0.158 & 498.5/88.1 \\
 & Random & 50.39 {\scriptsize $\pm$ 2.31} & 49.53 {\scriptsize $\pm$ 2.46} & 0.130 $\rightarrow$ 0.132 & 327.3/57.0 & 53.43 {\scriptsize $\pm$ 1.90} & 55.14 {\scriptsize $\pm$ 2.00} & 0.130 $\rightarrow$ 0.129 & 515.4/89.4 \\
\cmidrule(lr){2-10}
\multicolumn{2}{l|}{\textit{Ensemble Average}} & \textit{51.81 {\scriptsize $\pm$ 1.66}} & \textit{51.62 {\scriptsize $\pm$ 2.26}} & \textit{0.130 $\rightarrow$ 0.141} & \textit{309.2/54.6} & \textit{53.97 {\scriptsize $\pm$ 0.39}} & \textit{54.95 {\scriptsize $\pm$ 0.53}} & \textit{0.130 $\rightarrow$ 0.133} & \textit{510.6/89.0} \\
\bottomrule
\end{tabular}
}
\end{table*}

\subsection{Result details for DeepSeek-V3.2}

Tables~\ref{tab:exhaustive_infidabench_deepseekv32} and~\ref{tab:exhaustive_insightbench_deepseekv32} report the same exhaustive sweep for the DeepSeek-V3.2 suite. The qualitative pattern of the Haiku suite replicates --- Ensemble generation dominates on Infi-DA-Bench, no selection strategy separates from random, and the two aggregation arms track each other closely --- at a smaller budget return (${+}4.8$ vs.\ ${+}8.4$pp for $M{=}3{\to}5$ on Infi-DA-Bench), which is the model dependence noted in Section~\ref{sec:results}.

\begin{table*}[htb]
\centering
\caption{Results on \textbf{InFi-DA-Bench} with DeepSeek-V3.2 ($N{=}10$). We report self-aggregation and leader-aggregation accuracy (mean $\pm$ std), entropy for all plans $\rightarrow$ selected plans, and average token usage per question (in thousands) across three random seeds. Bold marks the best accuracy per column within each generation block. Both aggregation arms use identical candidate plans and differ only in the aggregating model, and on InFi-DA-Bench both arms are scored by the same evaluator, so the two accuracy columns are directly comparable. Entropy is computed from Titan embeddings of the plan texts.}
\label{tab:exhaustive_infidabench_deepseekv32}
\resizebox{\textwidth}{!}{
\begin{tabular}{ll|cccc|cccc}
\toprule
& & \multicolumn{4}{c|}{\textbf{M = 3}} & \multicolumn{4}{c}{\textbf{M = 5}} \\
\cmidrule(lr){3-6} \cmidrule(lr){7-10}
& & \multicolumn{2}{c}{\textbf{Accuracy (\%)}} & & & \multicolumn{2}{c}{\textbf{Accuracy (\%)}} & & \\
\cmidrule(lr){3-4} \cmidrule(lr){7-8}
\textbf{Generation} & \textbf{Selection} & \textbf{self-agg.} & \textbf{leader-agg.} & \textbf{Entropy} & \textbf{Tok In/Out (K)} & \textbf{self-agg.} & \textbf{leader-agg.} & \textbf{Entropy} & \textbf{Tok In/Out (K)} \\
\midrule
Base & Clustering & 72.11 {\scriptsize $\pm$ 2.87} & 72.50 {\scriptsize $\pm$ 1.57} & 0.118 $\rightarrow$ 0.143 & 121.1/12.2 & \textbf{80.29 {\scriptsize $\pm$ 0.98}} & \textbf{79.25 {\scriptsize $\pm$ 1.19}} & 0.119 $\rightarrow$ 0.138 & 157.7/17.3 \\
 & Goal Align & 73.15 {\scriptsize $\pm$ 1.70} & 72.11 {\scriptsize $\pm$ 2.38} & 0.118 $\rightarrow$ 0.121 & 120.6/12.1 & 78.08 {\scriptsize $\pm$ 1.47} & 77.69 {\scriptsize $\pm$ 2.87} & 0.118 $\rightarrow$ 0.116 & 156.4/17.2 \\
 & Maxent & \textbf{74.84 {\scriptsize $\pm$ 0.59}} & \textbf{73.41 {\scriptsize $\pm$ 0.98}} & 0.119 $\rightarrow$ 0.190 & 121.2/12.2 & 80.03 {\scriptsize $\pm$ 2.28} & 78.86 {\scriptsize $\pm$ 2.65} & 0.119 $\rightarrow$ 0.153 & 158.4/17.4 \\
 & Random & 74.19 {\scriptsize $\pm$ 0.22} & 73.15 {\scriptsize $\pm$ 0.00} & 0.120 $\rightarrow$ 0.118 & 120.2/12.0 & 77.04 {\scriptsize $\pm$ 2.17} & 76.39 {\scriptsize $\pm$ 0.81} & 0.129 $\rightarrow$ 0.128 & 165.0/17.8 \\
\cmidrule(lr){2-10}
\multicolumn{2}{l|}{\textit{Base Average}} & \textit{73.57 {\scriptsize $\pm$ 1.20}} & \textit{72.80 {\scriptsize $\pm$ 0.59}} & \textit{0.119 $\rightarrow$ 0.143} & \textit{120.8/12.1} & \textit{78.86 {\scriptsize $\pm$ 1.56}} & \textit{78.05 {\scriptsize $\pm$ 1.29}} & \textit{0.121 $\rightarrow$ 0.134} & \textit{159.4/17.4} \\
\midrule
Conditional & Clustering & 72.63 {\scriptsize $\pm$ 1.12} & 72.37 {\scriptsize $\pm$ 3.57} & 0.118 $\rightarrow$ 0.142 & 121.7/12.2 & \textbf{79.90 {\scriptsize $\pm$ 1.96}} & 78.21 {\scriptsize $\pm$ 1.70} & 0.118 $\rightarrow$ 0.138 & 157.3/17.3 \\
 & Goal Align & 73.28 {\scriptsize $\pm$ 2.28} & 72.63 {\scriptsize $\pm$ 2.81} & 0.119 $\rightarrow$ 0.118 & 121.7/12.2 & 77.56 {\scriptsize $\pm$ 0.45} & \textbf{78.47 {\scriptsize $\pm$ 1.37}} & 0.118 $\rightarrow$ 0.117 & 156.9/17.2 \\
 & Maxent & \textbf{74.06 {\scriptsize $\pm$ 0.98}} & \textbf{73.28 {\scriptsize $\pm$ 0.81}} & 0.119 $\rightarrow$ 0.191 & 121.7/12.2 & 79.51 {\scriptsize $\pm$ 0.59} & 77.30 {\scriptsize $\pm$ 1.57} & 0.117 $\rightarrow$ 0.151 & 157.0/17.3 \\
 & Random & 73.54 {\scriptsize $\pm$ 1.78} & 72.63 {\scriptsize $\pm$ 2.65} & 0.117 $\rightarrow$ 0.118 & 120.8/12.1 & 78.21 {\scriptsize $\pm$ 1.03} & 76.39 {\scriptsize $\pm$ 0.98} & 0.118 $\rightarrow$ 0.118 & 156.5/17.2 \\
\cmidrule(lr){2-10}
\multicolumn{2}{l|}{\textit{Conditional Average}} & \textit{73.38 {\scriptsize $\pm$ 0.59}} & \textit{72.73 {\scriptsize $\pm$ 0.39}} & \textit{0.118 $\rightarrow$ 0.142} & \textit{121.5/12.2} & \textit{78.79 {\scriptsize $\pm$ 1.09}} & \textit{77.59 {\scriptsize $\pm$ 0.94}} & \textit{0.118 $\rightarrow$ 0.131} & \textit{156.9/17.3} \\
\midrule
Ensemble & Clustering & 76.01 {\scriptsize $\pm$ 1.57} & 75.62 {\scriptsize $\pm$ 1.80} & 0.119 $\rightarrow$ 0.147 & 125.5/12.6 & 79.38 {\scriptsize $\pm$ 0.67} & 79.51 {\scriptsize $\pm$ 0.98} & 0.119 $\rightarrow$ 0.141 & 160.3/17.6 \\
 & Goal Align & \textbf{78.21 {\scriptsize $\pm$ 2.17}} & \textbf{76.78 {\scriptsize $\pm$ 2.14}} & 0.118 $\rightarrow$ 0.119 & 126.2/12.6 & 80.80 {\scriptsize $\pm$ 1.37} & 79.77 {\scriptsize $\pm$ 2.06} & 0.120 $\rightarrow$ 0.120 & 171.8/18.3 \\
 & Maxent & 76.01 {\scriptsize $\pm$ 0.59} & 76.52 {\scriptsize $\pm$ 1.37} & 0.116 $\rightarrow$ 0.187 & 124.8/12.5 & 80.29 {\scriptsize $\pm$ 0.98} & 79.38 {\scriptsize $\pm$ 0.67} & 0.119 $\rightarrow$ 0.153 & 160.9/17.7 \\
 & Random & 75.62 {\scriptsize $\pm$ 0.81} & 74.45 {\scriptsize $\pm$ 1.57} & 0.119 $\rightarrow$ 0.118 & 125.3/12.6 & \textbf{81.58 {\scriptsize $\pm$ 1.19}} & \textbf{81.58 {\scriptsize $\pm$ 0.59}} & 0.118 $\rightarrow$ 0.118 & 160.7/17.6 \\
\cmidrule(lr){2-10}
\multicolumn{2}{l|}{\textit{Ensemble Average}} & \textit{76.46 {\scriptsize $\pm$ 1.18}} & \textit{75.84 {\scriptsize $\pm$ 1.06}} & \textit{0.118 $\rightarrow$ 0.143} & \textit{125.4/12.6} & \textit{80.51 {\scriptsize $\pm$ 0.93}} & \textit{80.06 {\scriptsize $\pm$ 1.03}} & \textit{0.119 $\rightarrow$ 0.133} & \textit{163.4/17.8} \\
\bottomrule
\end{tabular}
}
\end{table*}

\begin{table*}[htb]
\centering
\caption{Results on \textbf{InsightBench} with DeepSeek-V3.2 ($N{=}10$). Accuracy is the proportion of the 428 ground-truth insights retrieved across the 100 questions, so one percentage point corresponds to $\approx$4.3 insights. We report self-aggregation and leader-aggregation accuracy (mean $\pm$ std), entropy for all plans $\rightarrow$ selected plans, and average token usage per question (in thousands) across three random seeds. Bold marks the best accuracy per column within each generation block. Both aggregation arms use identical candidate plans and differ only in the aggregating model, and on InsightBench both arms are scored by the same evaluator, so the two accuracy columns are directly comparable. Entropy is computed from Titan embeddings of the plan texts.}
\label{tab:exhaustive_insightbench_deepseekv32}
\resizebox{\textwidth}{!}{
\begin{tabular}{ll|cccc|cccc}
\toprule
& & \multicolumn{4}{c|}{\textbf{M = 3}} & \multicolumn{4}{c}{\textbf{M = 5}} \\
\cmidrule(lr){3-6} \cmidrule(lr){7-10}
& & \multicolumn{2}{c}{\textbf{Accuracy (\%)}} & & & \multicolumn{2}{c}{\textbf{Accuracy (\%)}} & & \\
\cmidrule(lr){3-4} \cmidrule(lr){7-8}
\textbf{Generation} & \textbf{Selection} & \textbf{self-agg.} & \textbf{leader-agg.} & \textbf{Entropy} & \textbf{Tok In/Out (K)} & \textbf{self-agg.} & \textbf{leader-agg.} & \textbf{Entropy} & \textbf{Tok In/Out (K)} \\
\midrule
Base & Clustering & 46.18 {\scriptsize $\pm$ 1.41} & 46.88 {\scriptsize $\pm$ 2.81} & 0.159 $\rightarrow$ 0.187 & 230.5/24.4 & 42.76 {\scriptsize $\pm$ 0.62} & 45.48 {\scriptsize $\pm$ 1.05} & 0.160 $\rightarrow$ 0.186 & 338.6/37.5 \\
 & Goal Align & 46.88 {\scriptsize $\pm$ 4.41} & 45.95 {\scriptsize $\pm$ 1.56} & 0.160 $\rightarrow$ 0.163 & 238.8/24.4 & 47.20 {\scriptsize $\pm$ 1.46} & 46.88 {\scriptsize $\pm$ 1.48} & 0.158 $\rightarrow$ 0.160 & 336.5/37.1 \\
 & Maxent & 44.94 {\scriptsize $\pm$ 2.01} & \textbf{47.35 {\scriptsize $\pm$ 1.35}} & 0.160 $\rightarrow$ 0.246 & 228.1/24.3 & 46.26 {\scriptsize $\pm$ 0.23} & \textbf{48.13 {\scriptsize $\pm$ 0.84}} & 0.159 $\rightarrow$ 0.202 & 347.2/38.1 \\
 & Random & \textbf{48.52 {\scriptsize $\pm$ 0.71}} & 46.18 {\scriptsize $\pm$ 2.39} & 0.162 $\rightarrow$ 0.162 & 228.4/24.5 & \textbf{47.66 {\scriptsize $\pm$ 3.85}} & 47.59 {\scriptsize $\pm$ 3.38} & 0.159 $\rightarrow$ 0.159 & 342.7/37.8 \\
\cmidrule(lr){2-10}
\multicolumn{2}{l|}{\textit{Base Average}} & \textit{46.63 {\scriptsize $\pm$ 1.49}} & \textit{46.59 {\scriptsize $\pm$ 0.64}} & \textit{0.160 $\rightarrow$ 0.189} & \textit{231.4/24.4} & \textit{45.97 {\scriptsize $\pm$ 2.22}} & \textit{47.02 {\scriptsize $\pm$ 1.15}} & \textit{0.159 $\rightarrow$ 0.176} & \textit{341.3/37.6} \\
\midrule
Conditional & Clustering & \textbf{49.38 {\scriptsize $\pm$ 3.78}} & 46.81 {\scriptsize $\pm$ 3.81} & 0.161 $\rightarrow$ 0.187 & 232.3/24.7 & 46.73 {\scriptsize $\pm$ 4.51} & 46.73 {\scriptsize $\pm$ 1.07} & 0.163 $\rightarrow$ 0.188 & 348.2/37.8 \\
 & Goal Align & 46.18 {\scriptsize $\pm$ 2.39} & 47.20 {\scriptsize $\pm$ 2.30} & 0.159 $\rightarrow$ 0.166 & 234.8/24.3 & \textbf{47.82 {\scriptsize $\pm$ 2.31}} & \textbf{47.43 {\scriptsize $\pm$ 1.53}} & 0.160 $\rightarrow$ 0.162 & 354.1/38.3 \\
 & Maxent & 45.17 {\scriptsize $\pm$ 1.87} & 42.99 {\scriptsize $\pm$ 1.64} & 0.159 $\rightarrow$ 0.245 & 234.5/24.8 & 45.64 {\scriptsize $\pm$ 1.72} & 45.72 {\scriptsize $\pm$ 2.45} & 0.161 $\rightarrow$ 0.204 & 354.1/37.6 \\
 & Random & 46.57 {\scriptsize $\pm$ 1.69} & \textbf{48.05 {\scriptsize $\pm$ 3.58}} & 0.161 $\rightarrow$ 0.168 & 223.7/23.8 & 45.64 {\scriptsize $\pm$ 3.11} & 46.57 {\scriptsize $\pm$ 2.35} & 0.158 $\rightarrow$ 0.158 & 340.6/37.5 \\
\cmidrule(lr){2-10}
\multicolumn{2}{l|}{\textit{Conditional Average}} & \textit{46.83 {\scriptsize $\pm$ 1.80}} & \textit{46.26 {\scriptsize $\pm$ 2.24}} & \textit{0.160 $\rightarrow$ 0.191} & \textit{231.3/24.4} & \textit{46.46 {\scriptsize $\pm$ 1.04}} & \textit{46.61 {\scriptsize $\pm$ 0.70}} & \textit{0.161 $\rightarrow$ 0.178} & \textit{349.3/37.8} \\
\midrule
Ensemble & Clustering & 45.17 {\scriptsize $\pm$ 1.99} & 44.78 {\scriptsize $\pm$ 0.88} & 0.166 $\rightarrow$ 0.193 & 234.9/25.3 & 49.84 {\scriptsize $\pm$ 2.48} & 48.36 {\scriptsize $\pm$ 1.76} & 0.170 $\rightarrow$ 0.194 & 348.9/38.3 \\
 & Goal Align & 46.57 {\scriptsize $\pm$ 3.42} & 44.39 {\scriptsize $\pm$ 0.23} & 0.168 $\rightarrow$ 0.133 & 236.4/25.1 & 46.88 {\scriptsize $\pm$ 3.40} & 48.13 {\scriptsize $\pm$ 2.23} & 0.169 $\rightarrow$ 0.148 & 357.5/38.9 \\
 & Maxent & \textbf{49.30 {\scriptsize $\pm$ 2.30}} & \textbf{47.98 {\scriptsize $\pm$ 1.95}} & 0.166 $\rightarrow$ 0.246 & 228.5/24.6 & \textbf{50.16 {\scriptsize $\pm$ 3.59}} & 47.59 {\scriptsize $\pm$ 0.88} & 0.169 $\rightarrow$ 0.209 & 350.5/38.5 \\
 & Random & 45.48 {\scriptsize $\pm$ 3.11} & 47.66 {\scriptsize $\pm$ 1.62} & 0.166 $\rightarrow$ 0.165 & 229.8/24.5 & 46.50 {\scriptsize $\pm$ 3.09} & \textbf{48.60 {\scriptsize $\pm$ 0.47}} & 0.168 $\rightarrow$ 0.169 & 352.7/38.8 \\
\cmidrule(lr){2-10}
\multicolumn{2}{l|}{\textit{Ensemble Average}} & \textit{46.63 {\scriptsize $\pm$ 1.88}} & \textit{46.20 {\scriptsize $\pm$ 1.88}} & \textit{0.167 $\rightarrow$ 0.184} & \textit{232.4/24.9} & \textit{48.35 {\scriptsize $\pm$ 1.92}} & \textit{48.17 {\scriptsize $\pm$ 0.43}} & \textit{0.169 $\rightarrow$ 0.180} & \textit{352.4/38.6} \\
\bottomrule
\end{tabular}
}
\end{table*}

\subsection{Reference configuration and full configuration-space ranking}
\label{app:refconfig}

Table~\ref{tab:config_space} extends the top-$3$ view of Table~\ref{tab:top_configs} to the complete $48$-point space. The reference configuration$^\dagger$ used throughout the paper (\texttt{Ensemble}+\texttt{Random}, $M{=}5$, leader aggregation) is the rank-$1$ row; it is adopted not because it is separably the most accurate --- five other rows, all \texttt{Ensemble} at $M{=}5$, are statistically indistinguishable from it under the paired bootstrap ($p>0.05$, up to $p=0.29$) --- but because it is the only member of that tied group whose selector adds zero overhead. Two structural reads of the full ranking corroborate the main-effects analysis: every $M{=}5$ \texttt{Ensemble} row outranks every non-\texttt{Ensemble} row at the same budget, while rows differing only in their selector are interleaved throughout; and P(best) concentrates entirely within the \texttt{Ensemble} block. One caution: P(best) measures the sampling stability of the argmax \emph{within} this setting, not transfer across settings --- restricting the same analysis to $M{=}3$ promotes \texttt{Ensemble}+\texttt{MaxEnt} to the top instead, a further symptom of the null selection axis. The generation choice, by contrast, is stable everywhere.


\begin{table}[t]
\centering
\caption{Complete ranking of the 48-point configuration space, with generation, selection, budget $M$ and aggregation arm all treated as design parameters. Accuracy is averaged over both base models, three seeds and the two benchmarks with equal weight; its absolute level is therefore not interpretable, while comparisons between rows are, since every row is evaluated on the same questions. \emph{P(best)} is the share of $10000$ paired bootstrap resamples in which a row attains the highest overall score, resampling question clusters once per replicate and rescoring every configuration on that same resample; chance is $2.1\%$. Shading marks the first (\colorbox{goldrow}{gold}), second (\colorbox{silverrow}{silver}) and third (\colorbox{bronzerow}{bronze}) ranked configurations. }
\label{tab:config_space}
\footnotesize
\setlength{\tabcolsep}{4pt}
\begin{tabular}{rllclrrrr}
\toprule
& \multicolumn{4}{l}{\textbf{Configuration}} & \multicolumn{3}{c}{\textbf{Accuracy (\%)}} & \\
\cmidrule(lr){2-5} \cmidrule(lr){6-8}
\textbf{\#} & \textbf{Gen.} & \textbf{Sel.} & $\mathbf{M}$ & \textbf{Agg.} & \textbf{Infi} & \textbf{Ins.} & \textbf{Overall} & \textbf{P(best)} \\
\midrule
\rowcolor{goldrow}
1 & Ensemble & Random & 5 & leader & 82.49 & 51.87 & \textbf{67.18} & 57.7\% \\
\rowcolor{silverrow}
2 & Ensemble & Goal-Align & 5 & leader & 82.36 & 51.21 & 66.78 & 20.8\% \\
\rowcolor{bronzerow}
3 & Ensemble & Goal-Align & 5 & self & 82.49 & 50.55 & 66.52 & 9.6\% \\
4 & Ensemble & MaxEnt & 5 & leader & 80.87 & 51.56 & 66.21 & 3.9\% \\
5 & Ensemble & Clustering & 5 & leader & 80.80 & 51.60 & 66.20 & 3.5\% \\
6 & Ensemble & MaxEnt & 5 & self & 79.96 & 52.22 & 66.09 & 3.1\% \\
7 & Ensemble & Random & 5 & self & 81.65 & 49.96 & 65.80 & 0.6\% \\
8 & Ensemble & Clustering & 5 & self & 79.25 & 51.91 & 65.58 & 0.7\% \\
9 & Base & MaxEnt & 5 & leader & 76.52 & 52.14 & 64.33 & $<$0.1\% \\
10 & Ensemble & MaxEnt & 3 & leader & 76.98 & 51.40 & 64.19 & $<$0.1\% \\
11 & Base & Random & 5 & leader & 76.52 & 51.25 & 63.89 & $<$0.1\% \\
12 & Conditional & Goal-Align & 5 & leader & 77.04 & 50.55 & 63.79 & $<$0.1\% \\
13 & Ensemble & MaxEnt & 3 & self & 75.81 & 51.71 & 63.76 & $<$0.1\% \\
14 & Conditional & MaxEnt & 5 & leader & 76.46 & 50.78 & 63.62 & $<$0.1\% \\
15 & Conditional & Clustering & 5 & self & 77.17 & 49.96 & 63.57 & $<$0.1\% \\
16 & Conditional & Random & 5 & leader & 75.81 & 51.13 & 63.47 & $<$0.1\% \\
17 & Conditional & Clustering & 5 & leader & 77.50 & 49.34 & 63.42 & $<$0.1\% \\
18 & Base & Clustering & 5 & leader & 77.56 & 49.03 & 63.29 & $<$0.1\% \\
19 & Ensemble & Goal-Align & 3 & self & 77.50 & 48.71 & 63.11 & $<$0.1\% \\
20 & Base & MaxEnt & 5 & self & 76.01 & 50.16 & 63.08 & $<$0.1\% \\
21 & Ensemble & Goal-Align & 3 & leader & 78.02 & 47.78 & 62.90 & $<$0.1\% \\
22 & Conditional & MaxEnt & 5 & self & 76.46 & 49.26 & 62.86 & $<$0.1\% \\
23 & Base & Goal-Align & 5 & leader & 76.07 & 49.49 & 62.78 & $<$0.1\% \\
24 & Conditional & Goal-Align & 5 & self & 74.32 & 50.97 & 62.65 & $<$0.1\% \\
25 & Conditional & Random & 5 & self & 75.62 & 49.45 & 62.54 & $<$0.1\% \\
26 & Ensemble & Clustering & 3 & self & 76.46 & 48.52 & 62.49 & $<$0.1\% \\
27 & Base & Random & 5 & self & 74.64 & 50.31 & 62.48 & $<$0.1\% \\
28 & Ensemble & Clustering & 3 & leader & 77.04 & 47.86 & 62.45 & $<$0.1\% \\
29 & Base & Goal-Align & 5 & self & 74.84 & 49.81 & 62.32 & $<$0.1\% \\
30 & Ensemble & Random & 3 & leader & 74.84 & 48.60 & 61.72 & $<$0.1\% \\
31 & Base & Clustering & 5 & self & 76.33 & 47.08 & 61.70 & $<$0.1\% \\
32 & Ensemble & Random & 3 & self & 74.97 & 47.94 & 61.45 & $<$0.1\% \\
33 & Conditional & Clustering & 3 & self & 69.26 & 51.52 & 60.39 & $<$0.1\% \\
34 & Base & MaxEnt & 3 & leader & 69.97 & 50.47 & 60.22 & $<$0.1\% \\
35 & Conditional & Clustering & 3 & leader & 70.17 & 50.23 & 60.20 & $<$0.1\% \\
36 & Conditional & Random & 3 & leader & 69.71 & 50.55 & 60.13 & $<$0.1\% \\
37 & Base & Clustering & 3 & leader & 68.74 & 50.58 & 59.66 & $<$0.1\% \\
38 & Base & MaxEnt & 3 & self & 69.84 & 49.14 & 59.49 & $<$0.1\% \\
39 & Base & Random & 3 & self & 68.74 & 50.12 & 59.43 & $<$0.1\% \\
40 & Conditional & Goal-Align & 3 & self & 69.07 & 49.65 & 59.36 & $<$0.1\% \\
41 & Conditional & Random & 3 & self & 69.78 & 48.71 & 59.25 & $<$0.1\% \\
42 & Base & Goal-Align & 3 & leader & 68.42 & 50.00 & 59.21 & $<$0.1\% \\
43 & Base & Goal-Align & 3 & self & 67.96 & 50.27 & 59.12 & $<$0.1\% \\
44 & Conditional & Goal-Align & 3 & leader & 68.74 & 49.10 & 58.92 & $<$0.1\% \\
45 & Base & Clustering & 3 & self & 67.64 & 50.16 & 58.90 & $<$0.1\% \\
46 & Conditional & MaxEnt & 3 & self & 68.68 & 48.99 & 58.83 & $<$0.1\% \\
47 & Base & Random & 3 & leader & 68.68 & 48.71 & 58.70 & $<$0.1\% \\
48 & Conditional & MaxEnt & 3 & leader & 69.65 & 47.51 & 58.58 & $<$0.1\% \\
\bottomrule
\end{tabular}
\end{table}

\section{Example tasks}

The two benchmarks jointly cover the enterprise analytics spectrum, illustrated by one task of each type.

\begin{figure}
    \centering
    \includegraphics[width=0.99\linewidth]{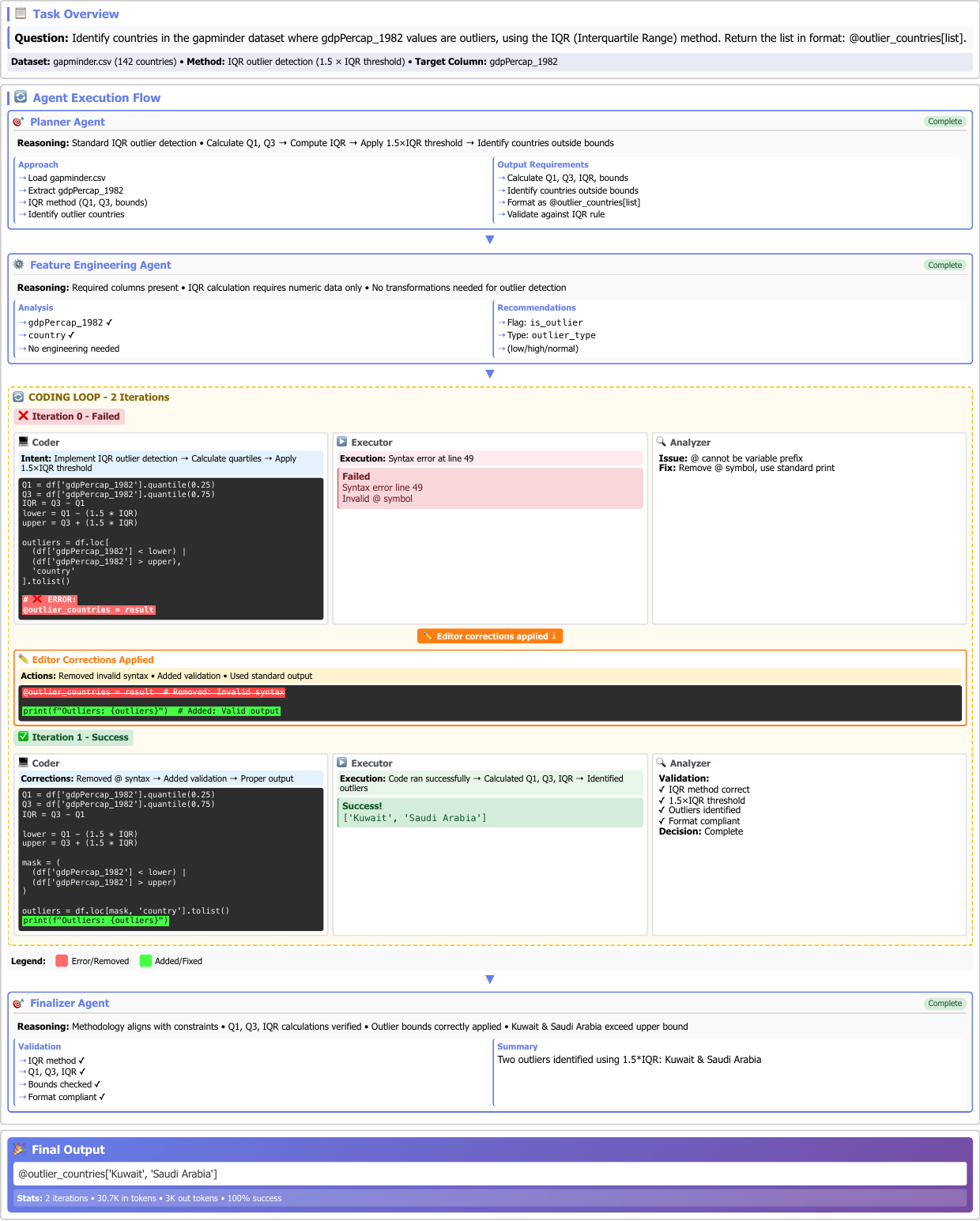}
    \caption{Example execution of CIPHER on question 254 of Infi-DA-Bench.}
    \label{fig:example_infidabench}
\end{figure}

\begin{figure}
    \centering
    \includegraphics[width=0.99\linewidth]{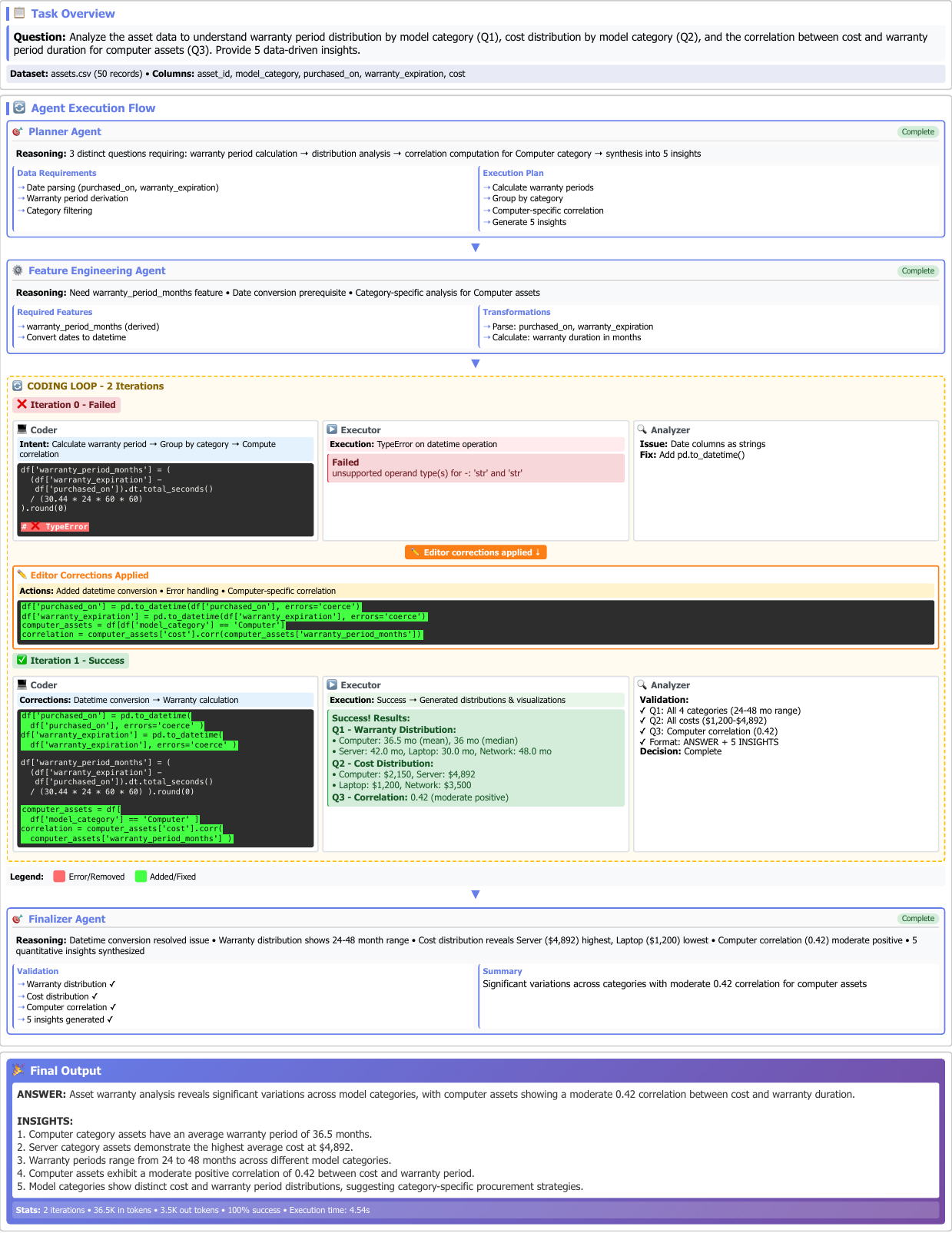}
    \caption{Example execution of CIPHER on question 16 of Insightbench.}
    \label{fig:example_insightbench}
\end{figure}

\section{Discussion and limitations}
\label{sec:discussion}

We formulate test-time scaling as a decoupled framework between generation and selection. The DES framework draws structural inspiration from the classic exploration-exploitation trade-off in reinforcement learning \citep{lattimore2020bandit}. By separating candidate generation ($N$) from state execution ($M$), it maps wide-breadth search space exploration, to a distinct phase of strategic exploitation. 
The proposed agent \texttt{CIPHER} performs competitively against state-of-the-art purpose-built specialist baselines on enterprise data analytics benchmarks. 

Our broad empirical study on the different designs of DES shows that generation and aggregation are the more effective levers, compared to the selection strategy. We show that the framework allows meaningful performance gains against standard test-time approaches (\texttt{Base} generation + \texttt{Random} selection). For practitioners provisioning inference compute, the same results read as an allocation rule: spend the budget on ensemble generation and a larger execution budget $M$, upgrade only the aggregator node when a stronger model is affordable, and spend less on plan selection.

\textbf{Limitations} Our selection strategies are agnostic to the token-level probabilities of the underlying LMs; some open-access models expose logits which could enable finer selection. \texttt{CIPHER} generates all $N$ candidate states in a single upfront batch, whereas adaptive schemes that grow the candidate pool in response to execution failures could reduce cost further. Our validation is focused on data science tasks; whether the DES framework transfers to other agent domains remains an open question.

\end{document}